\documentclass{article}

\usepackage{arxiv}
\usepackage{xcolor}
\usepackage{caption}
\usepackage{subcaption}
\usepackage{mathtools}
\usepackage{float}
\usepackage{booktabs,siunitx}
\usepackage{makecell}

\usepackage[utf8]{inputenc} 

\usepackage[T1]{fontenc}    
\usepackage{hyperref}       
\usepackage{url}            
\usepackage{booktabs}  
\usepackage{amsfonts}       
\usepackage{amssymb}
\usepackage{amsthm}
\usepackage{nicefrac}       
\usepackage{microtype}      
\usepackage{lipsum}
\usepackage{graphicx}
\usepackage{cite}
\usepackage{amsmath}
\usepackage{color}
\usepackage{cancel}
\usepackage{caption}
\usepackage{subcaption}
\newcommand{\rr}{\color{red}}

\DeclareMathOperator{\sech}{sech}

\title{TanhSoft - a family of activation functions combining Tanh and Softplus}

\author{
  Koushik Biswas\\
  Department of Computer Science \\
  IIIT Delhi\\
  New Delhi, India, 110020 \\
  \texttt{koushikb@iiitd.ac.in} \\
   \And
 Sandeep Kumar \\
  Department of Computer Science, IIIT Delhi\\ \& \\
  Department of Mathematics, Shaheed Bhagat Singh College,\\ University of Delhi.\\ 
  New Delhi, India. \\
  \texttt{sandeepk@iiitd.ac.in, sandeep\_kumar@sbs.du.ac.in} \\
   \And
 Shilpak Banerjee \\
  Department of Mathematics \\
  IIIT Delhi\\
  New Delhi, India, 110020 \\
  \texttt{shilpak@iiitd.ac.in} \\
   \And
 Ashish Kumar Pandey \\
  Department of Mathematics\\
  IIIT Delhi\\
  New Delhi, India, 110020 \\
  \texttt{ashish.pandey@iiitd.ac.in} \\
}
\date{}

\begin{document}
\maketitle

\begin{abstract}

Deep learning at its core, contains functions that are composition of a linear transformation with a non-linear function known as activation function. In past few years, there is an increasing interest in construction of novel activation functions resulting in better learning. In this work, we propose a family of novel activation functions, namely {\em TanhSoft}, with four undetermined hyper-parameters of the form $\tanh{(\alpha x+ \beta e^{\gamma x})} \ln(\delta+e^x)$ and tune these hyper-parameters to obtain activation functions which are shown to outperform several well known activation functions. For instance, replacing ReLU with $x \tanh(0.6e^x)$ improves top-1 classification accuracy on CIFAR-10 by 0.46\% for DenseNet-169 and 0.7\% for Inception-v3 while with $\tanh(0.87x)\ln(1+e^x)$ top-1 classification accuracy on CIFAR-100 improves by 1.24\% for DenseNet-169 and 2.57\% for SimpleNet model. 
\end{abstract}


\section{Introduction}
Artificial neural networks (ANNs) have occupied the center stage in the realm of deep learning in the recent past. ANNs are made up of several hidden layers, and each hidden layer consists of several neurons. At each neuron, an affine linear map is composed with a non-linear function known as \textit{activation function}. During the training of an ANN, the linear map is optimized, however an activation function is usually fixed in the beginning along with the architecture of the ANN. There has been an increasing interest in developing a methodical understanding of activation functions, in particular with regards to the construction of novel activation functions and identifying mathematical properties leading to a better learning\cite{survey}. 

An activation function is considered good if it can increase the learning rate and leaning to better convergence which leads to more accurate results. At the early stage of deep learning research, researchers used shallow networks (fewer hidden layers), and tanh or sigmoid, were used as activation functions. As the research progressed and deeper networks (multiple hidden layers) came into fashion to achieve challenging tasks, Rectified Linear Unit (ReLU)(\cite{relu},\cite{relu1},\cite{relu2}) emerged as the most popular activation function. 
Despite its simplicity, deep neural networks with ReLU have learned many complex and highly nonlinear functions with high accuracy.

To overcome the shortcomings of ReLU (non-zero mean, negative missing, unbounded output, to name a few, see\cite{srs} and to increase the accuracy considerably in a variety of tasks in comparison to networks with ReLU, many new activation functions have been proposed over the years. Many of these new activation functions are variants of ReLU, for example, 
Leaky ReLU \cite{lReLU}, Exponential Linear Unit (ELU) \cite{elu}, Parametric Rectified Linear Unit (PReLU) \cite{pReLU}, Randomized Leaky Rectified Linear Units (RReLU) \cite{rReLU} and Inverse Square Root Linear Units (ISRLUS) \cite{isReLU}. In the recent past, some activation functions constructed from tanh or sigmoid have achieved state-of-the-art results on a variety of challenging datasets. Most notably, among such activation functions, Swish \cite{swish} has emerged as a close favorite to ReLU. Some of these novel activation functions have shown that introducing of hyper-parameters in the argument of the functions may provide activation functions for special values of these hyper-parameters that can outperform activation functions for other values of hyper-parameters, for example, see\cite{swish}, \cite{srs}.  



In this article, we propose a family of activation functions with four hyper-parameters of the form 
\begin{align}
    f(x;\alpha,\beta,\gamma,\delta) = \tanh(\alpha x+ \beta e^{\gamma x})\ln(\delta+e^x).
\end{align}
We show that activation functions for some specific values of hyper-parameters outperform several well known and conventional activation functions, including ReLU and Swish. Moreover, using a hyper-parameterized combination of known activation functions, we attempt to make the search for novel activation functions organized. As indicated above and validated below, such an organized search can often find better performing activation functions in the vicinity of known functions. 

\section{Related works}

\begin{figure}[H]
    \begin{minipage}[t]{.49\linewidth}
        \centering
    
        \includegraphics[width=\linewidth]{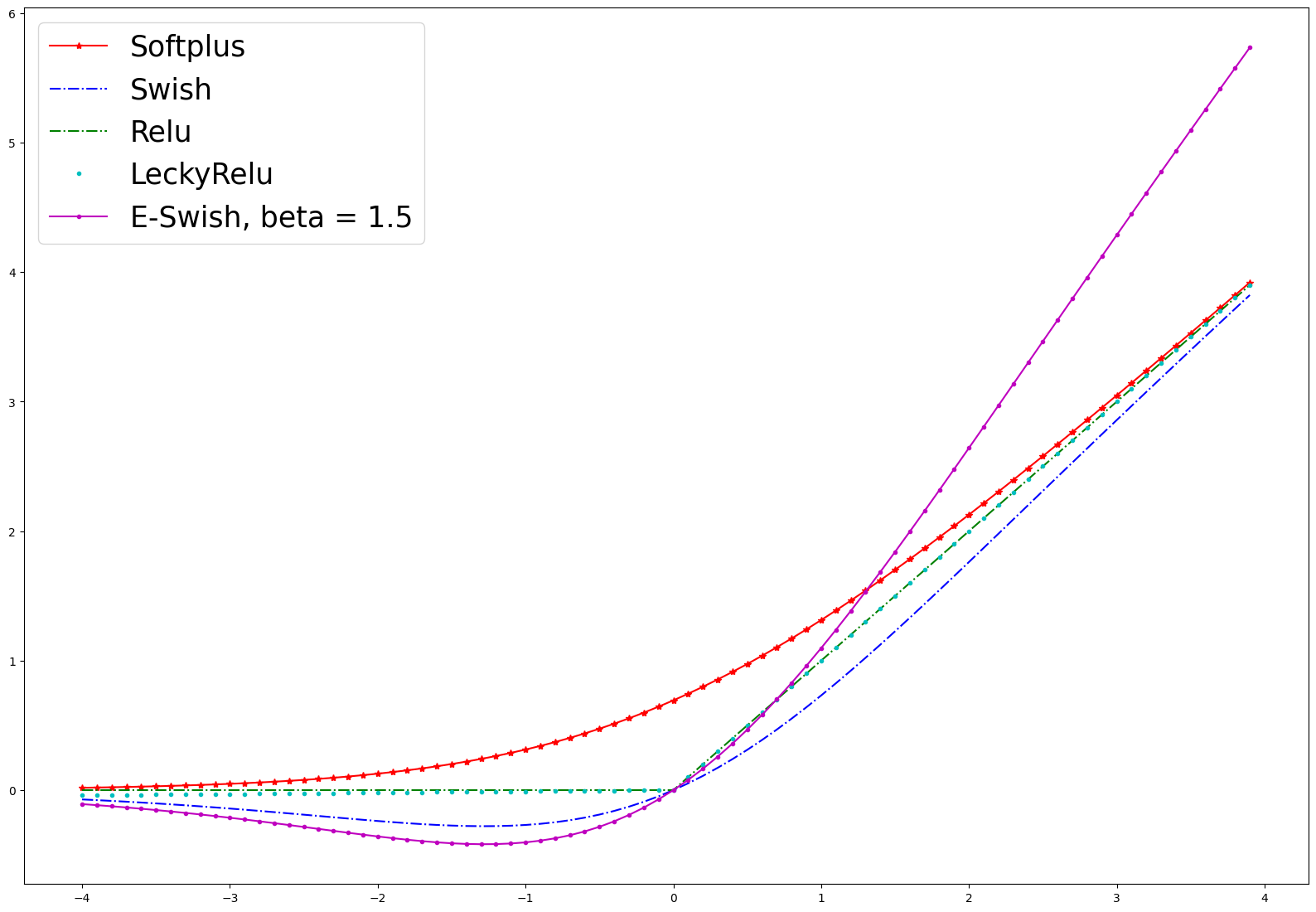}
        
        \label{fig1}
    \end{minipage}
    \hfill
    \begin{minipage}[t]{.49\linewidth}
        \centering
        
       \includegraphics[width=\linewidth]{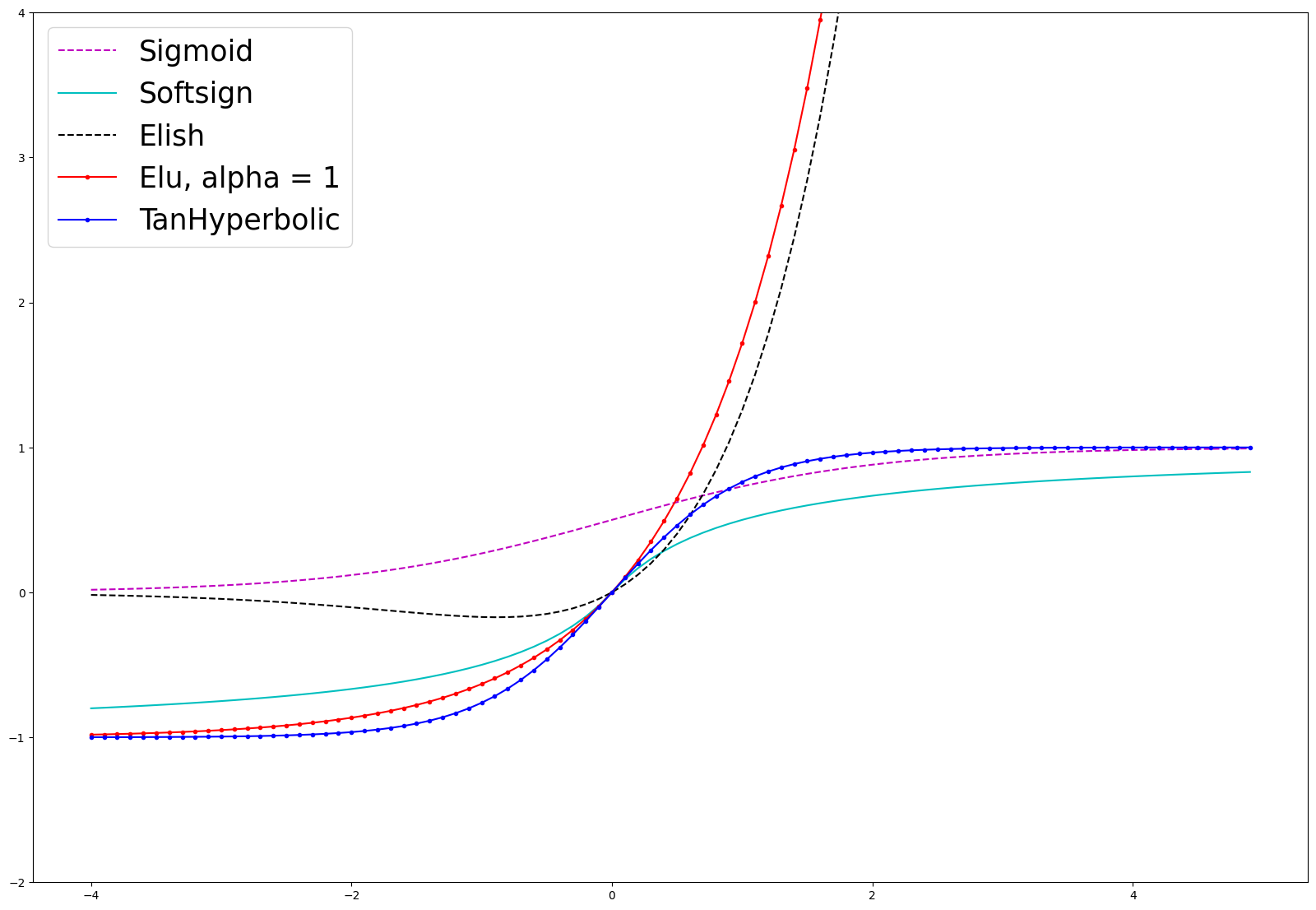}
        
        \label{fig1a}
    \end{minipage}  
    \caption{Plot of various activation functions}
\label{all}
\end{figure}

We give a brief idea of a few most widely used activation functions. All of these functions, along with some members of TanhSoft family, are given in Figure~\ref{all}.
\begin{itemize}

\item \textbf{Sigmoid:-}
Sigmoid activation function, which is also known as logistic function, used in binary classification problem in outcome layers and it produces outputs based on probability. Sigmoid is a smooth, bounded, non-linear and differentiable function range in $(0,1)$. Sigmoid suffers from vanishing gradient problem. Sigmoid is defined as
\begin{align}
    \sigma(x) = \frac{1}{1+e^{-x}}
\end{align}

\item \textbf{Hyperbolic Tangent Function:-}
Hyperbolic Tangent Function, tanh, is a smooth, non-linear and differentiable function in the range $(-1,1)$ defined as
\begin{align}
    \tanh(x) = \frac{e^x-e^{-x}}{e^x+e^{-x}}.
\end{align}
It is used in recurrent neural networks, natural language processing\cite{tanh} and speech processing tasks but also suffers from vanishing gradient problem.

\item \textbf{ Rectified Linear Unit (ReLU):-}
The rectified linear unit (ReLU) activation function was first introduced by Nair and Hinton in 2010 [REF]. At present, It is one of the most widely used activation function. ReLU is a kind of linear function, and it is identity in the positive axis while $0$ in the negative axis. One of the best property about ReLU is, it learns really fast. ReLU suffers from vanishing gradient problem. Also, in some situation, it has been observed that up-to 50$\%$ of neurons are dead because of 0 value in the negative axis. ReLU(\cite{relu2}, \cite{relu1}, \cite{relu}) is defined as 
\begin{align}
    f(x) = \max(0,x).
\end{align}
\item \textbf{Leaky Rectified Linear Unit:-}
Leaky Rectified Linear Unit(Leaky ReLU) was proposed by Mass et al. on 2013\cite{lReLU}. Leaky ReLU has introduced an non-zero gradient in the negative axis to overcome the vanishing gradient and dead neuron problems of ReLU. LReLU is defined as 
\begin{align} 
f(x)& = 
    \begin{cases}
       x & x > 0 \\               
      0.01.x & x\leq 0
    \end{cases}
\end{align}
\item \textbf{Parametric ReLU:-} 
Parametric ReLU(PReLU) is similar to Leaky ReLU where Leaky ReLU has predetermined negative slope where PReLU has a parametric negative slope. PReLU is defined as 
\begin{align} 
f(x)& = 
    \begin{cases}
       x & x > 0 \\               
      a.x & x\leq 0
    \end{cases}
\end{align}
where $a\leq 1$ and so f(x) is equivalent to max(x,ax).

\item \textbf{Swish:-}
To construct Swish, researcher from Google brain team uses reinforcement learning based automatic search technique. It was proposed by Ramachandran et al., on 2017\cite{swish}. Swish is a non-monotonic, smooth function which is bounded below and unbounded above. Swish is defined as 
\begin{align}
    f(x) = x \sigma(x) = \frac{x}{1+e^{-x}}
\end{align}
\item \textbf{E-swish:-}
E-Swish \cite{eswish} is a slight modified version of Swish function, introduced by Alcaide on 2018, defined by
\begin{align}
    f(x) = \beta x \sigma(x)
\end{align} where $\beta$ is a trainable parameter. This function follows all the properties of Swish and can provide better results than Swish as claimed in the paper for some values of $\beta$.

\item \textbf{ELISH:-}
Exponential Linear Sigmoid SquasHing(ELiSH) was proposed by Barisat et el. \cite{elish} on 2018. It is unbounded above, bounded below, non-monotonic, and
smooth function defined by
\begin{align} 
f(x)& = 
    \begin{cases}
       \frac{x}{1+e^{-x}} & x \geq 0 \\               
      \frac{e^x-1}{1+e^{-x}} & x < 0
    \end{cases}
\end{align}

\item \textbf{Softsign:-}
 The Softsign function was proposed by Turian et al., 2009 \cite{softsign}. Softsign is a quadratic polynomial based function. Softsign is used in regression problem \cite{reg} as well as speech system \cite{speech} and ac hived some promising results. Softsign is defined as
 \begin{align}
     f(x) = \frac{x}{1+|x|}
 \end{align}
 where $|x|$ is absolute value of x.\\

\item \textbf{Exponential Linear Units:-}
Exponential Linear Units(ELU) was proposed by Clevert et al., 2015 \cite{elu}. ELU is deifined in such a way so that it overcomes the vanishing gradient problem of ReLU. ELU is a fast learner and it generalises better than ReLU and LReLU. ELU is defined as 
\begin{align} 
f(x)& = 
    \begin{cases}
       x & x > 0 \\               
      \alpha (e^x-1) & x\leq 0
    \end{cases}
\end{align}
where $\alpha$ is a hyper-parameter. 

\item \textbf{Softplus:-}
Softplus  was  proposed  by  Dugas  et  al., 2001 \cite{softplus, softplus1}. Softplus is a smooth activation function and has non-zero gradient. Softplus is defined as 
\begin{align}
    \operatorname{Softplus}(x) = ln(1+e^x)
\end{align}

\end{itemize}


\section{TanhSoft activation function family}
Standard ANN training process involves tuning the weights in the linear part of the network, however there is a merit in the ability to custom design activation functions, to better fit the problem at hand. Here, rather than looking at individual activation functions, we  propose a family of functions indexed by four hyper-parameters. We refer to this family as the TanhSoft family as it is created by combining a hyper-parametric versions of the $\tanh$ and the Softplus activation functions. Explicitly, we express it as
\begin{align}\label{eq: tanhsoft family}
    f(x;\alpha,\beta,\gamma,\delta) = \tanh(\alpha x+ \beta e^{\gamma x})\ln(\delta+e^x).
\end{align}
Any function in this family can be used as a activation function for hyper-parameter values $-\infty\leq \alpha\leq 1$ , $0\leq \beta < \infty, 0< \gamma < \infty$ and $0\leq\delta\leq 1$, though  for experimental and practical purpose, we restrict to small ranges of the hyper-parameters ($0\leq \alpha\leq 3 , 0\leq \beta < 2, 0< \gamma < 4$ and $\delta = 0,1$). The interplay between the hyper-parameters $\alpha, \beta$, and  $\gamma$ plays a major role for the TanhSoft family and controls the slope of the curve in both positive and negative axes. The hyper-parameter $\delta$ is used as a switch to change the SoftPlus component of TanhSoft to a linear function ($\delta=0$) allowing us to cover a larger class of functions. 

Note that $f(x; 0,0,\gamma,\delta)$ recovers the zero function and $f(x; 0,\beta,0,0)$ becomes the linear function family $\tanh(\beta)x$. For large values of some parameter while fixing the other parameters, the TanhSoft family converges to some known activation functions pointwise. For example,
\begin{align*}
    \lim_{\gamma\to\infty}f(x;0, \beta, \gamma, 0)= \operatorname{ReLU}(x) \quad\forall x\in\mathbb{R}\text{ for any fixed }\beta>0.
\end{align*}
With the similar hyper-parametric settings except for very small negative values of $\alpha$, $f(x; \alpha,\beta,\gamma,0)$ has behaviour similar to the Parametric ReLU activation function. 
Also,
\begin{align*}
    \lim_{\beta\to\infty}f(x;0, \beta, \gamma, 1)= \operatorname{Softplus}(x) \quad\forall x\in\mathbb{R}\text{ for any fixed }\gamma>0.
\end{align*}
We remark that the MISH\cite{mish} activation function is also obtained from $\tanh$ and Softplus but it is a composition while we use a hyper-parametric product. Worth noting that the author has reported unstable training behaviour for the specific function $f(x;1,0,\gamma,1)$ in \cite{mish}, however, we failed to find any instability during the training process. Also, in \cite{tanhexp} the authors have mentioned the function $f(x;0,1,1,0)$, which arise as an example from the TanhSoft family. In fact, we show that because of the introduction of hyper-parameters, better activation functions of the form $f(x;0,\beta,\gamma,0)$ can be obtained.

Being a product of two smooth functions, TanhSoft is a family of smooth activation functions. As expected, the monotonocity and boundedness of the functions in the family depend on the specific values of the hyper-parameters. The derivative of the TanhSoft family is given by
\begin{align}\label{eq: tanhsoft family derivative}
    f'(x; \alpha,\beta,\gamma, \delta)= \tanh(\alpha x+ \beta e^{\gamma x})\frac{e^x}{\delta+e^x}+(\alpha+\beta \gamma e^{\gamma x})\sech^2(\alpha x+\beta e^{\gamma x})\ln(\delta+e^x).
\end{align}
A detailed study of the mathematical properties of the TanhSoft family will be presented in a later work. In this work, we focus on providing several examples of activation functions from the family which perform well on many challenging datasets.

\section{Search Findings}
We have performed an organized search of activation functions within the TanhSoft family by varying the values of hyper-parameters and training and testing them with DenseNet-121\cite{densenet} and SimpleNet\cite{simple} on CIFAR10\cite{cifar10} dataset. Several functions were tested and we select eight functions as example to report their performance.
The Top-1 and Top-3 accuracies of these eight functions are given in Table~\ref{tab112}. All these functions, either outperformed or give near accuracy as compared to ReLU. Most notably, $f(x;0,0.6,1,0) = x\tanh(0.6e^x)$ and $f(x;0.87,0,\gamma,1)= \tanh(0.87x)\ln(1+e^x)$ constantly outperform ReLU even with more complex models. We have given detailed results in Section~\ref{sec:exp} with more complex models and datasets.

\begin{figure}[H]
    \begin{minipage}[t]{.49\linewidth}
        \centering
    
        \includegraphics[width=\linewidth]{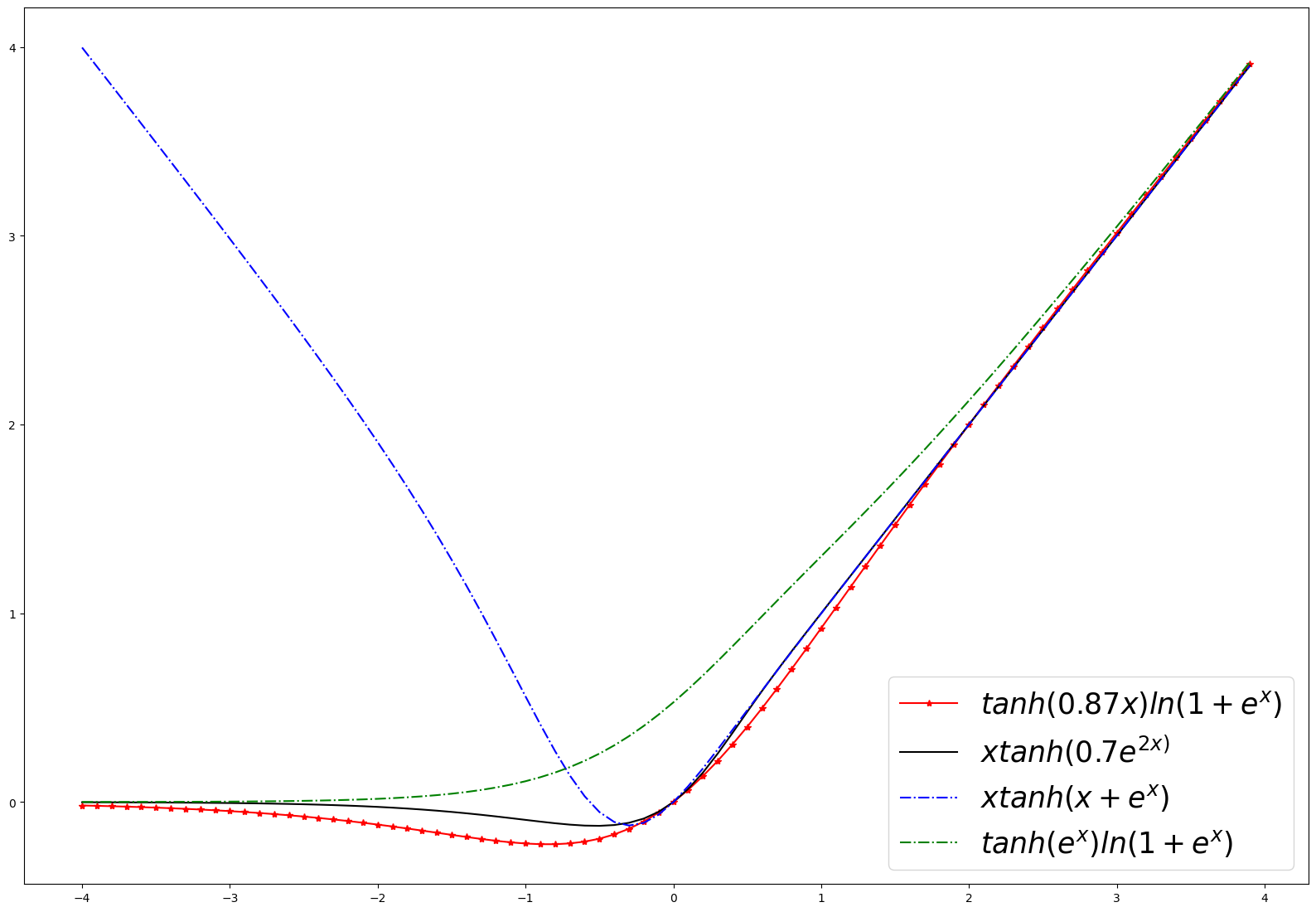}
        
        \label{ser1}
    \end{minipage}
    \hfill
    \begin{minipage}[t]{.49\linewidth}
        \centering
        
       \includegraphics[width=\linewidth]{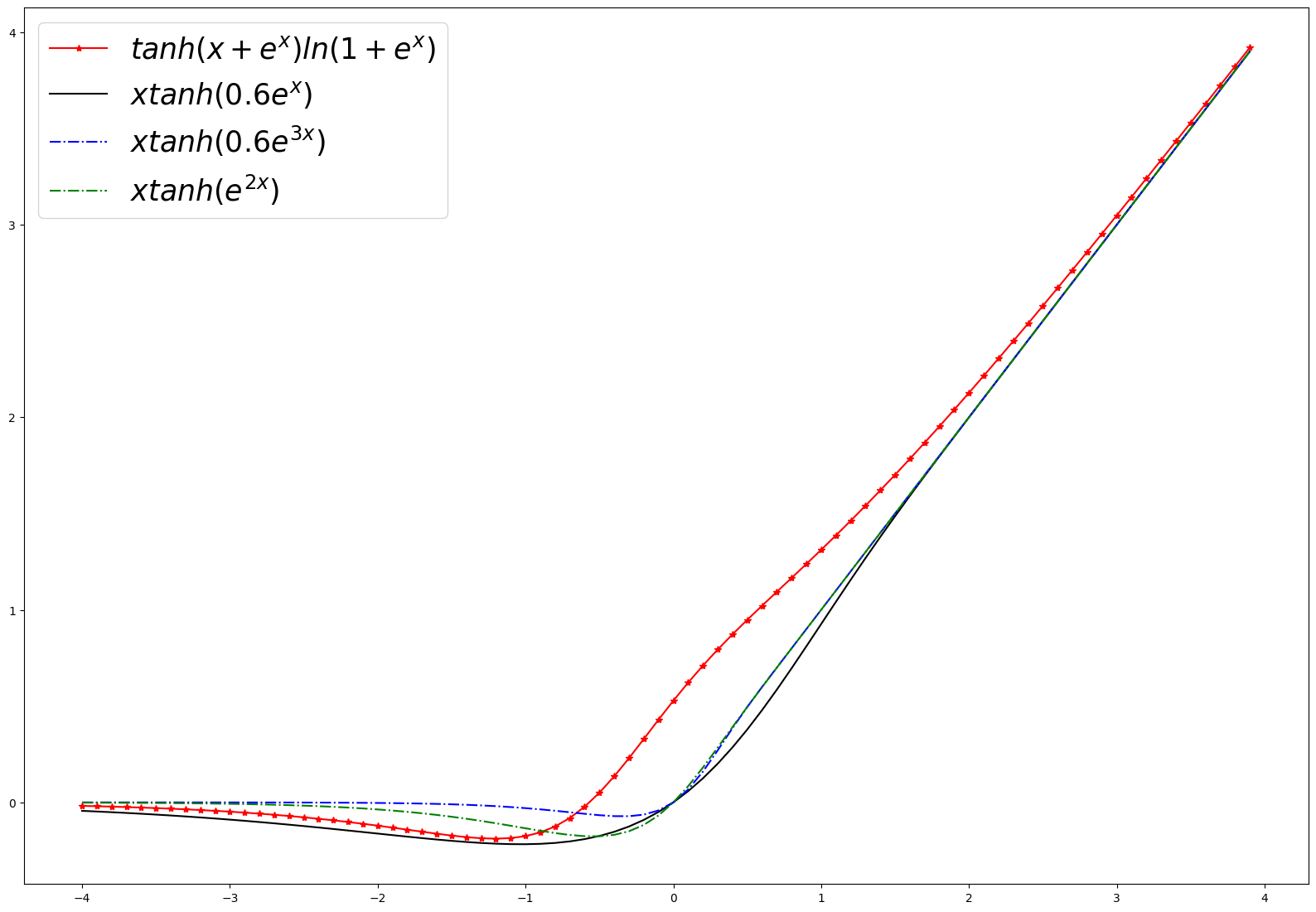}
        
        \label{fig1a}
    \end{minipage}  
    \caption{ A few novel activation functions from the searches of the TanhSoft family.}
\label{ser2}
\end{figure}

\begin{table}[H]
\newenvironment{amazingtabular}{\begin{tabular}{*{50}{l}}}{\end{tabular}}
\centering
\begin{amazingtabular}
\midrule
\makecell{Activation \\ Function} & \makecell{Top-1 accuracy on \\ DenseNet-121} & \makecell{Top-3 accuracy on\\ DenseNet-121} & \makecell{Accuracy on \\ SimpleNet Model}\\
\midrule
$\operatorname{ReLU}(x)=\max(0,x)$ & \hspace{0.9cm}90.73 & \hspace{0.9cm}98.73 & \hspace{0.9cm}91.01\\
\midrule
\midrule
$f(x;0,0.6,1,0) = x\tanh(0.6e^x)$ & \hspace{0.9cm}91.14 & \hspace{0.9cm}98.86 & \hspace{0.9cm}92.23\\
\midrule
$f(x;0.87,0,\gamma,1) = \tanh(0.87x)\ln(1+e^x)$ & \hspace{0.9cm}90.98 & \hspace{0.9cm}98.80 & \hspace{0.9cm}92.07\\
\midrule
$f(x;0,1,2,0) = x\tanh(e^{2x})$ &  \hspace{0.9cm}90.84 & \hspace{0.9cm}98.64 & \hspace{0.9cm}91.38\\
\midrule
$f(x;0,0.6,3,0) = x\tanh(0.6e^{3x})$ &  \hspace{0.9cm}90.60 & \hspace{0.9cm}98.80  & \hspace{0.9cm}91.20\\
\midrule
$f(x;0,1,1,1) = \tanh(e^x)\ln(1+e^x)$ & \hspace{0.9cm}90.69 & \hspace{0.9cm}98.42 & \hspace{0.9cm}91.35\\
\midrule
$f(x;0,0.7,2,0) = x\tanh(0.7e^{2x})$ & \hspace{0.9cm}91.11 & \hspace{0.9cm}98.66 & \hspace{0.9cm}91.98 \\
\midrule
$f(x;1,1,1,0) = x\tanh(x+e^x)$ & \hspace{0.9cm}90.78 & \hspace{0.9cm}98.67 & \hspace{0.9cm}89.42 \\
\midrule
$f(x;1,1,1,1) = \tanh(x+e^x)\ln(1+e^x)$ & \hspace{0.9cm}90.43 & \hspace{0.9cm}98.62 & \hspace{0.9cm}91.58\\
\midrule

\end{amazingtabular}
\vspace{0.5cm}
  \caption{Accuracy on CIFAR-10 for eight example functions from the TanhSoft family along with ReLU.}
  \label{tab112}
\end{table}
Real-world datasets are noisy or challenging, and it is always difficult to find the best activation function to generalize on random datasets. It is hard to say whether the searched function will generalize successfully and replace ReLU on challenging or noisy datasets. Though there may be merit in having a custom activation function corresponding to the problem at hand, but yet it is beneficial to identify activation functions that generalize to several real world data sets, making it easier to implement. Hence we concentrate on two members from the TanhSoft family and establish their generalizability and usefulness over other conventional activation functions. In particular, we consider the sub-families  $f(x;\alpha,0,\gamma,1)$ and $f(x;0,\beta,\gamma,0)$ and call them as \textit{TanhSoft-1} and \textit{TanhSoft-2}. In what follows, we discuss the properties of these subfamilies, experiments with more complex models, and comparison with a few other widely used activation functions.
\section{TanhSoft-1 and TanhSoft-2}
The functions, TanhSoft-1 and TanhSoft-2 are given as
\begin{align}
    & \text{TanhSoft-1}:\qquad f(x;\alpha,0,\gamma,1) = \tanh(\alpha x) \text{Softplus}(x) = \tanh(\alpha x) \ln(1+e^x),\\
    & \text{TanhSoft-2}:\qquad f(x;0,\beta,\gamma,0) = x\tanh(\beta e^{\gamma x}).
\end{align}
The corresponding derivatives (see equation \ref{eq: tanhsoft family derivative}) are
\begin{align}
   & \text{Derivative \ of \ TanhSoft-1}:\qquad  f'(x;\alpha ,0,\gamma ,1) = \tanh(\alpha x) \frac{ e^x}{(1+e^x)}+\alpha \sech^2(\alpha x) \ln(1+e^x),\\
   & \text{Derivative \ of \ TanhSoft-2}:\qquad f'(x;0,\beta ,\gamma ,0) = \tanh(\beta   e^{\gamma x}) + \beta \gamma x e^{ \gamma x} \sech^2(\beta e^{\gamma x}).
\end{align}

\begin{figure}[H]
    \begin{minipage}[t]{.47\linewidth}
        \centering
    
        \includegraphics[width=\linewidth]{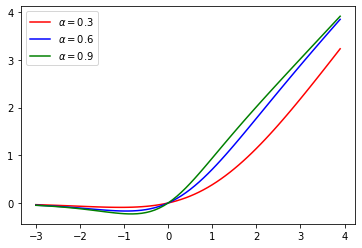}
        \caption{TanhSoft-1 Activation for different values of $\alpha$}
        \label{fig11}
    \end{minipage}
    \hfill
    \begin{minipage}[t]{.49\linewidth}
        \centering
       \includegraphics[width=\linewidth]{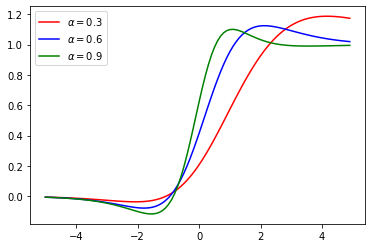}
        \caption{First order derivative Derivative of TanhSoft-1 Activation for different values of $\alpha$}
        \label{fig1a}
    \end{minipage}  
\end{figure}

\begin{figure}[H]
    \begin{minipage}[t]{.47\linewidth}
        \centering
    
        \includegraphics[width=\linewidth]{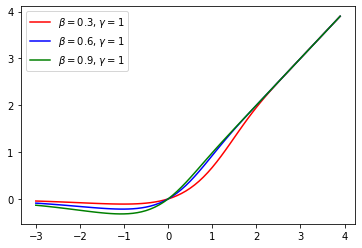}
        \caption{TanhSoft-2 Activation for different values of $\beta$}
        \label{fig12}
    \end{minipage}
    \hfill
    \begin{minipage}[t]{.49\linewidth}
        \centering
        
       \includegraphics[width=\linewidth]{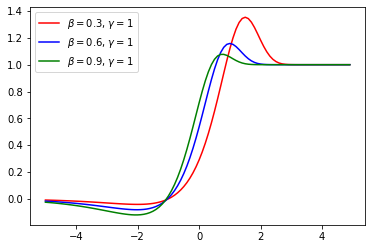}
        \caption{First order derivative Derivative of TanhSoft-2 Activation for different values of $\beta$}
        \label{fig12a}
    \end{minipage}  
\end{figure}

\begin{figure}[H]
    \begin{minipage}[t]{.47\linewidth}
        \centering
    
        \includegraphics[width=\linewidth]{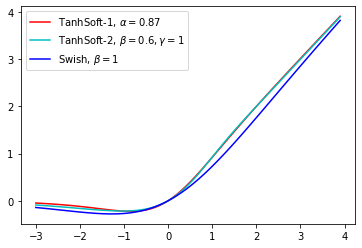}
        \caption{TanhSoft-1, TanhSoft-2 and Swish}
        \label{com}
    \end{minipage}
    \hfill
    \begin{minipage}[t]{.49\linewidth}
        \centering
        
       \includegraphics[width=\linewidth]{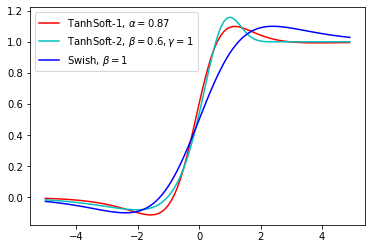}
        \caption{First order derivatives of TanhSoft-1, TanhSoft-2 and Swish}
        \label{deri}
    \end{minipage}  
\end{figure}

Figure~\ref{fig11} and ~\ref{fig12} show the graph for TanhSoft-1 and TanhSoft-2 activation functions for different values of $\alpha$, and $\beta$ and $\gamma$ respectively. If $\alpha = 0$, then TanhSoft-1 becomes the zero function. Similarly, for $\beta = 0$, TanhSoft-2 is the zero function. Like ReLU and Swish, TanhSoft-1 and TanhSoft-2  are also unbounded above but bounded below. Like Swish, TanhSoft-1 and TanhSoft-2 are both smooth, non-monotonic activation functions.
Plots of the first derivative of TanhSoft-1 and TanhSoft-2 are given in Figures~\ref{fig1a} and ~\ref{fig12a} for different values of $\alpha$, and $\beta$ and $\gamma$ respectively. A comparison between TanhSoft-1, TanhSoft-2 and Swish and their first derivatives are given in Figures~\ref{com} and \ref{deri}.

TanhSoft-1 and TanhSoft-2 can be implemented with a single line of code in Keras Library \cite{keras} or Tensorflow V2.3.0 \cite{tensorflow}. The Keras codes for TanhSoft-1 and TanhSoft-2 are
\begin{align*}
& \text{tf.keras.activations.tanh}(\alpha * x) * \text{tf.keras.activations.softplus}(x) \qquad\text{ and, }\\
& x* \text{tf.keras.activations.tanh}(\beta * \text{tf.keras.activations.exp}(\gamma x))
\end{align*} for specific value of $\alpha, \beta$ and $\gamma$ respectively.


\subsection*{Experiments with TanhSoft-1 and TanhSoft-2} \label{sec:exp}
We tested TanhSoft-1 and TanhSoft-2 for different values of hyper-parameters against widely used activation functions on CIFAR and MNIST datasets. In particular, TanhSoft-1 with $\alpha=0.87$, $f(x;0.87,0,\gamma,1)$, and TanhSoft-2  with $\beta=0.6, \gamma=1$, $f(x;0,0.6,1,0)$, produced best results. We observe that $f(x;0.87,0,\gamma,1)$ and $f(x;0,0.6,1,0)$ in most cases beats or performs equally with baseline activation functions, and underperforms marginally on rare occasions. Table~\ref{tab49} gives a comparison with the baseline activation such as ReLU, Leaky ReLU, ELU, Softplus, and Swish. We have tested our activation function in several models, such as Densenet121\cite{densenet}, DenseNet169\cite{densenet}, InceptionNet V3\cite{incep}, SimpleNet\cite{simple}, MobileNets\cite{mobile}, WideResNet 28-10\cite{wrn}. In this next section we will provide details of our experimental framework and results. 

\begin{table}[H]
\newenvironment{amazingtabular}{\begin{tabular}{*{50}{l}}}{\end{tabular}}
\centering
\begin{amazingtabular}
\midrule
Baselines & ReLU & Leaky ReLU & ELU & Swish & Softplus\\
\midrule
TanhSoft-1 > Baseline & \hspace{0.3cm}10 & \hspace{0.45cm}7 & \hspace{0.3cm}11 & \hspace{0.3cm}10 & \hspace{0.3cm}10\\
TanhSoft-1 = Baseline & \hspace{0.3cm}1 & \hspace{0.45cm}2 & \hspace{0.3cm}0 & \hspace{0.3cm}1 & \hspace{0.3cm}0\\
TanhSoft-1 < Baseline & \hspace{0.3cm}0 & \hspace{0.45cm}2 & \hspace{0.3cm}0 & \hspace{0.3cm}0 & \hspace{0.3cm}1\\
\midrule
TanhSoft-2 > Baseline & \hspace{0.3cm}11 & \hspace{0.45cm}9 & \hspace{0.3cm}11 & \hspace{0.3cm}11 & \hspace{0.3cm}11\\
TanhSoft-2 = Baseline & \hspace{0.3cm}0 & \hspace{0.45cm}1 & \hspace{0.3cm}0 & \hspace{0.3cm}0 & \hspace{0.3cm}0\\
TanhSoft-2 < Baseline & \hspace{0.3cm}0 & \hspace{0.45cm}1 & \hspace{0.3cm}0 & \hspace{0.3cm}0 & \hspace{0.3cm}0\\
\midrule
\end{amazingtabular}
\vspace{0.5cm}
  \caption{Baseline table of TanhSoft-1 and TanhSoft-2 for Top-1 Accurecy}
  \label{tab49}
\end{table}
\subsubsection*{MNIST}
MNIST \cite{mnist} database contains image data of handwritten digits from 0 to 9. The dataset contains 60k training and 10k testing $28\times 28$ grey-scale images. We run a custom 5-layer CNN architecture on MNIST dataset with different activation functions and results are reported in Table~\ref{tab1}. We have reported accuracy and loss for TanhSoft-1 for different values of $\alpha$ in Figures~\ref{acc} and \ref{loss}.

\begin{table}[H]
\begin{center}
\begin{tabular}{ |c|c|c| }
 \hline
 Activation Function &  5-fold Accuracy on MNIST data  \\
 \hline
 TanhSoft-1($\alpha = 0.87$) &  99.0 ($\pm$ 0.05)\\ 
 \hline
 TanhSoft-2($\beta = 0.6, \gamma=1$) &  \textbf{99.1 ($\pm$ 0.05)}\\ 
 \hline
 ReLU  &  99.0 ($\pm$ 0.06) \\ 
 \hline
 Swish  & 98.9 ($\pm$ 0.08) \\
 \hline
 Leaky ReLU($\alpha$ = 0.01) & 99.0 ($\pm$ 0.1)  \\
 \hline
 ELU  & 98.9($\pm$ 0.06) \\
 \hline
 Softplus & 98.9($\pm$ 0.06)\\
 \hline
 \end{tabular}
 \vspace{0.2cm}
\caption{Experimental Results with MNIST Dataset} 
\label{tab1}
\end{center}
\end{table}


\begin{figure}[H]
    \begin{minipage}[t]{.48\linewidth}
        \centering
    
        \includegraphics[width=\linewidth]{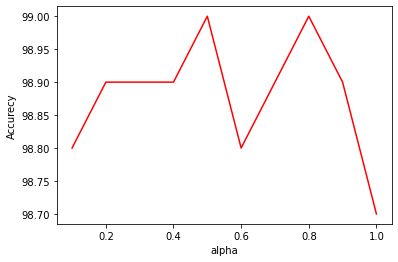}
        \caption{Accuracy with TanhSoft-1 Activation function in MNIST dataset for different values for $\alpha$}
        \label{acc}
    \end{minipage}
    \hfill
    \begin{minipage}[t]{.49\linewidth}
        \centering
        
       \includegraphics[width=\linewidth]{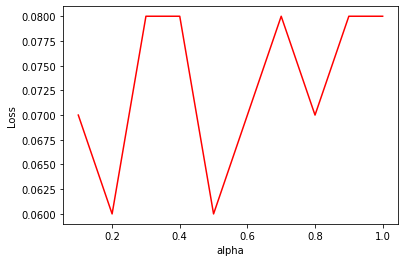}
        \caption{Loss with TanhSoft-1 Activation function in MNIST dataset for different values of $\alpha$}
        \label{loss}
    \end{minipage}  
\end{figure}


\subsubsection*{CIFAR}
The CIFAR \cite{cifar10} dataset consists of $32\times 32$ colored images, consists of 60k images and divided into 50k training and 10k test images. There are two type of CIFAR dataset such as CIFAR10 and CIFAR100. CIFAR10 dataset has 10 classes with 6000 images per class while for CIFAR100 has 100 classes with 600 images per class. We have reported the results of TanhSoft-1 and TanhSoft-1 for $\alpha = 0.87$ and $\beta = 0.6, \gamma = 1$ respectively along with ReLU, Leaky ReLu, ELU, Softplus and Swish in the CIFAR10 dataset with DenseNet121, DenseNet169, InceptionNet V3 and SimpleNet while for CIFAR100 dataset results have been reported with DenseNet121, DenseNet169, InceptionNet V3, MobileNet\cite{mobile}, WideResNet\cite{wrn} and SimpleNet. We have trained with Adam optimizer\cite{adam} with 100 epochs for DenseNet121, DenseNet169, InceptionNet V3, MobileNet, WideResNet and 200 epochs for SimpleNet. We used  weight decay $5\times 10^{-4}$ in SimpleNet model. Weight decay has been decided according to\cite{weight}. ~Table \ref{tab2}, \ref{tab3} contains results for CIFAR10 data while ~Table \ref{tab26}, \ref{tab27} contains results for CIFAR100 data. ~Table \ref{tab45} contains results for accuracy and Loss on CIFAR10 dataset with SimpleNet Model and TanhSoft-2 activation function for different values of $\beta$ and $\gamma = 1$


\begin{table}[H]
  \centering
  \begin{tabular}{@{}rrrrrrrrrrr@{}}
    \multicolumn{1}{c}{} & \multicolumn{3}{c}{} \\
   
    & \multicolumn{1}{c}{} & \multicolumn{1}{c}{} & \multicolumn{1}{c}{}
    & \multicolumn{1}{c}{}
    & \multicolumn{1}{c}{}
    & \multicolumn{1}{c}{}
    & \multicolumn{1}{c}{}
    & \multicolumn{1}{c}{}
    & \multicolumn{1}{c}{}
    & \multicolumn{1}{c}{}\\
    \cmidrule{2-11}
    \vspace{0.3cm}
    $\beta$ & 0.1 & 0.2 & 0.3 & 0.4 & 0.5 & 0.6 & 0.7 & 0.8 & 0.9 & 1.0\\
    \cmidrule{2-11}
    \vspace{0.3cm}
    Accuracy & 91.05 & 91.60 & 92.01 & 91.75 & 91.77 & 92.23 & 91.79 & 91.61 & 92.05 & 91.87 \\
    \cmidrule{2-11}
    \vspace{0.3cm}
    Loss & 0.5103 & 0.4631 & 0.4428 & 0.4659 & 0.4480 & 0.4204 & 0.4606 & 0.4688 & 0.4511 & 0.4578\\
    \cmidrule{2-11}
    
  \end{tabular}
  \vspace{0.5cm}
  \caption{Accuracy and Loss On CIFAR10 dataset with SimpleNet Model and TanhSoft-2 activation function for different values of $\beta$ and $\gamma = 1$}
  \label{tab45}
\end{table}

\begin{table}[!htbp]
\begin{center}
\begin{tabular}{ |c|c|c|c|c|c| }
 \hline
\makecell{Activation\\ Function} &  \makecell{DenseNet\\
121 Model\\
Top-1 \\ Accuracy}
& \makecell{DenseNet\\
121 Model\\ Top-3 \\ Accuracy}
& \makecell{DenseNet\\
169 Model\\
Top-1\\ Accuracy}
& \makecell{DenseNet\\
169 Model\\ Top-3\\ Accuracy}
\\ 
 \hline
 TanhSoft-1($\alpha = 0.87$) & 90.98   & 98.80  & 91.05  & 98.75\\ 
 \hline
 TanhSoft-2($\beta = 0.6, \gamma=1$) & \textbf{91.14}   & 98.86  & \textbf{91.10}  & 98.78\\
 \hline 
 ReLU  &  90.73 & 98.73 & 90.64 & 98.79 \\ 
 \hline
 Leaky ReLU($\alpha$ = 0.01) & 90.77 & 98.80 & 90.61 & 98.78  \\
 \hline
 ELU  & 90.49 & 98.61 & 90.40 & 98.65 \\
 \hline
 Swish & 90.77 & 98.80 & 90.38 & 98.68\\
 \hline
 Softplus & 90.45 & 98.65 & 90.51 & 98.69\\
 \hline
  
 \end{tabular}
 \vspace{0.2cm}
\caption{Experimental Results with CIFAR10 Dataset with DenseNet-121 and DenseNet-169} 
\label{tab2}
\end{center}
\end{table}
\begin{table}[!h]
\begin{center}
\begin{tabular}{ |c|c|c|c|c|}
 \hline
 \makecell{Activation\\ Function} & \makecell{SimpleNet\\ Top-1 Accuracy}
& 
\makecell{Inception-Net\\
Model V3\\
Top-1 \\Accuracy}
& \makecell{Inception\\
Model V3\\ Top-3\\ Accuracy}\\
\hline
TanhSoft-1($\alpha = 0.87$) & 92.07   & 91.93 & 98.84\\ 
 \hline
 TanhSoft-2($\beta = 0.6,\gamma=1$) & \textbf{92.23}   & \textbf{91.99} & 98.85\\
 \hline
 ReLU  &  91.01 & 91.29 & 98.80 \\ 
 \hline
 Leaky ReLU($\alpha$ = 0.01) & 91.05 & 91.84 & 98.93  \\
 \hline
 ELU  & 91.19 & 91.01 & 98.79 \\
 \hline
 Swish  &  91.59 & 91.26 & 98.75\\
 \hline
 Softplus  &  91.42 & 91.79 & 98.84\\
 \hline
\end{tabular}
\vspace{0.2cm}
\caption{Experimental Results with CIFAR10 Dataset with SimpleNet and Inception V3} 
\label{tab3}
\end{center}
\end{table}


\begin{figure}[!h]
    \begin{minipage}[t]{.485\linewidth}
        \centering
    
        \includegraphics[width=\linewidth]{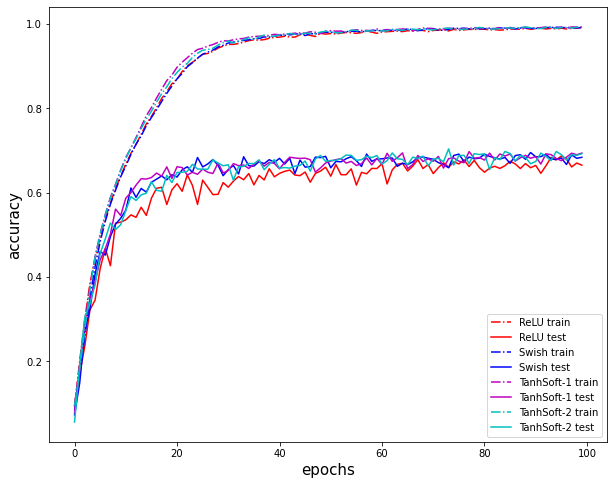}
        \caption{Train and Test accuracy on CIFAR100 dataset with WideResNet 28-10 model}
        \label{acc}
    \end{minipage}
    \hfill
    \begin{minipage}[t]{.48\linewidth}
        \centering
        
       \includegraphics[width=\linewidth]{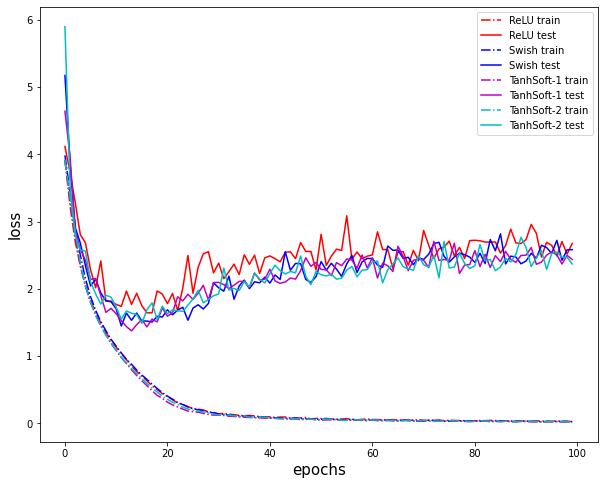}
        \caption{Train and Test loss on CIFAR100 dataset with WideResNet 28-10 model}
        \label{loss}
    \end{minipage}  
\end{figure}


\begin{table}[!h]
\begin{center}
\begin{tabular}{ |c|c|c|c|c|c| }
 \hline
\makecell{Activation\\ Function} &  \makecell{DenseNet\\
121 Model\\
Top-1 \\ Accuracy}
& \makecell{DenseNet\\
121 Model\\ Top-3 \\ Accuracy}
& \makecell{DenseNet\\
169 Model\\
Top-1\\ Accuracy}
& \makecell{DenseNet\\
169 Model\\ Top-3\\ Accuracy} & \makecell{SimpleNet \\ Top-1 Accuracy}
\\ 
 \hline
 TanhSoft-1($\alpha = 0.87$) &  66.99  &  83.76 & \textbf{65.37}  & 82.42 & \textbf{65.20}\\ 
 \hline
 TanhSoft-2($\beta = 0.6,\gamma=1$) &  \textbf{67.18}  &  84.01 & 64.99  & 82.09 & 65.01\\
 \hline
 ReLU  &  66.40 & 83.11 & 64.15 & 81.69 & 62.63\\ 
 \hline
 
 Leaky ReLU($\alpha$ = 0.01) & \textbf{67.18} & 83.37 & 63.40 & 81.16 & 62.58\\
 \hline
 ELU  & 66.52 & 83.42 & 64.23 & 81.45 & 63.74\\
 \hline
 Swish  &  66.99 & 83.76 & 64.95 & 82.06 & 64.90\\
 \hline
 Softplus  &  65.93 & 83.50 & 64.95 & 82.05 & 62.39\\
 \hline
  
 \end{tabular}
 \vspace{0.2cm}
\caption{Experimental Results with CIFAR100 Dataset with DenseNet and SimpleNet models} 
\label{tab26}
\end{center}
\end{table}

\begin{table}[H]
\begin{center}
\begin{tabular}{ |c|c|c|c|c|c| }
 \hline
\makecell{Activation\\ Function} &  \makecell{MobileNet\\
Top-1 \\ Accuracy}
& \makecell{MobileNet\\ Top-3 \\ Accuracy}
& \makecell{Inception V3\\
Top-1\\ Accuracy}
& \makecell{Inception V3\\ Top-3\\ Accuracy} & \makecell{WideResNet 28-10\\ Top-1 Accuracy}
\\ 
 \hline
 TanhSoft-1($\alpha = 0.87$) &  57.56  & 76.57 & 69.19 & 84.63 & 69.40\\ 
 \hline
 TanhSoft-2($\beta = 0.6$) &  57.56  &  76.57 &  \textbf{69.28} & 85.91 & \textbf{69.41}\\
 \hline 
 ReLU  & 56.87 & 76.33 & 69.09  & 85.41 & 66.54\\ 
 \hline
 Leaky ReLU($\alpha$ = 0.01) & \textbf{57.78} & 77.32 & 69.19 & 85.24 & 69.23\\
 \hline
 ELU  & 57.19 & 76.03 & 68.32 &  85.30 & 64.48\\
 \hline
 Swish & 55.40 & 74.81 & 67.61 & 83.59 & 68.45\\
 \hline
 SoftPlus & 57.53 & 76.48 & 69.24 & 85.15 & 61.53\\
 \hline
  
 \end{tabular}
 \vspace{0.2cm}
\caption{Experimental Results with CIFAR100 Dataset with MobileNet, Inception V3 and WideResNet 28-10} 
\label{tab27}
\end{center}
\end{table}

\section{Conclusion}
In this study, we have explored a new novel hyper-parameter family of activation functions, TanhSoft, defined mathematically as $\tanh{(\alpha x+ \beta e^{\gamma x})} \ln(\delta+e^x)$ where $\alpha$, $\beta$, $\gamma$ and $\delta$ are tunable hyper-parameters. We have shown that in the different complex models, TanhSoft outperforms in the MNIST, Cipher10, and CIFAR100 datasets compared to ReLU, Leaky ReLU, Swish, ELU, and Softplus so that TanhSoft can be a good choice to replace the  ReLU, Swish, and other widely used activation functions.
Future work can be, applying the proposed novel activation function to more challenging datasets such as ImageNet \& COCO and try on different other models to achieve  State-of-the-Art results.


\bibliographystyle{unsrt}  
\bibliography{main}





\end{document}